# Modeling Freight Mode Choice Using Machine Learning Classifiers: A Comparative Study Using the Commodity Flow Survey (CFS) Data

**Majbah Uddin**
Postdoctoral Research Associate
National Transportation Research Center
Oak Ridge National Laboratory
Knoxville, TN 37923
ORCiD: 0000-0001-9925-3881
Email: uddinm@ornl.gov

**Sabreena Anowar***
Assistant Professor
Department of Civil and Environmental Engineering and Department of Architectural Studies
University of Missouri – Columbia
Columbia, MO 65211
Email: anowars@missouri.edu
ORCiD: 0000-0002-1915-9594

**Naveen Eluru**
Associate Professor
Department of Civil, Environmental and Construction Engineering
University of Central Florida
Orlando, FL 32816
ORCiD number: 0000-0003-1221-4113
Email: naveen.eluru@ucf.edu

* Corresponding author

**Abstract**
This study aims to explore the usefulness of machine learning classifiers for modeling freight mode choice. We investigate eight commonly used machine learning classifiers, namely Naïve Bayes, Support Vector Machine, Artificial Neural Network, K-Nearest Neighbors, Classification and Regression Tree, Random Forest, Boosting and Bagging, along with the classical Multinomial Logit model. The 2012 Commodity Flow Survey data is used as the primary data source; we augment it with spatial attributes from secondary data sources. The performance of the classifiers is compared based on prediction accuracy results. The current research also examines the role of sample size and training-testing data split ratios on the predictive ability of the various approaches. In addition, the importance of variables is estimated to determine how the variables influence freight mode choice. The results show that the tree-based ensemble classifiers perform the best. Specifically, Random Forest produces the most accurate predictions, closely followed by Boosting and Bagging. With regard to variable importance, shipment characteristics, such as shipment distance, industry classification of the shipper and shipment size, are the most significant factor for freight mode choice decisions.

**Introduction**
Freight industry encompasses a significant component of the US economy. In fact, the value of annual freight shipments in US amount to upwards of 17 trillion dollars (Freight Analysis Framework, 2019). Given the magnitude of freight industry technological transformations underway in transportation and change in behavioral patterns of consumption in the US (and in general) could have significant impacts. In terms of technological transformation, the emergence of connected and autonomous vehicles can have significant implications for freight movement. While level 4 adoption which is a fully self-driving vehicle in all conditions, (as defined by NHTSA) is likely to take time, several intermediate levels of vehicle technologies are already being introduced by private and public companies. These vehicular advances offer significant advantages to the trucking industry in terms of fuel, time and labor cost savings. For instance, adoption of fully autonomous vehicles will allow the trucking industry to circumvent the need for federally mandated driver breaks for long-haul trips. Moreover, with the prevalence of online shopping (such as Amazon, Walmart) and crowd sourcing shipping options (such as Instacart, Doordash), the traditional pattern of freight flows is rapidly changing; particularly, the shipment size distribution is moving towards a higher share of smaller size shipments (Golob and Regan, 2001; Rotem-Mindali and Weltevreden, 2013). According to CFS data, in 2012, almost 90 percent commodities shipped were under 500 pounds and worth 25 percent by shipment value ($). A quantitative analysis for understanding the impact of these emerging changes is critical to proactively adjust/accommodate for the impacts on transportation industry and economy in general.

Of the different choice dimensions associated with freight flows, the mode(s) chosen for shipping freight is an important component of supply chain network and has significant implications for the transportation system and the environment at large. For instance, in addition to GHG emissions, movement of trucks is associated with negative externalities such as traffic congestion (ensuing delays), traffic crashes (ensuing property damage, injuries and fatalities) (Huang et al., 2010), and expedient transportation infrastructure (roadway and bridge surfaces) deterioration (Ponnuswamy and Johnson Victor, 2012). Moreover, the "unpriced" external cost (per ton-mile) of shipping freight using truck mode is eight times higher than the "unpriced" external cost of using rail mode (Austin, 2015). Given the wide-ranging implications, researchers have started to focus their attention towards developing freight mode choice models. A comprehensive understanding of the decision process involved in shipping freight by various modes would benefit transportation infrastructure planning decisions and their management.

The research on modeling freight mode choice has typically focused on the random utility based multinomial logit (MNL) model and its variants (such as nested logit, mixed MNL, and latent segmentation based MNL) (see Keya et al., 2017 for a detailed review). Over the last decade, driven by the enhancements in computing power and the advent of big data analytics, machine learning approaches have gained attention in the transportation research community. These methods include Naïve Bayes (NB), Support Vector Machine (SVM), Artificial Neural Networks (ANN), Classification and Regression Tree (CART), Random Forest (RF), Boosting and Bagging. With their inherent strength in handling large datasets, these approaches are well suited to extracting patterns that are often hard to accommodate within traditional econometric models.

To be sure, several research efforts have accommodated for machine learning approaches for modeling mode choice. The adoption of these approaches is more common in examining passenger mode choice analysis. A brief summary of earlier research on the use of machine learning (ML) methods for both passenger and freight mode choice are presented in Table 1. The information in the table includes study area, data source, modes considered, classical and ML methods used, comparison metrics, training-testing data split and major findings obtained from the comparison exercise with the classical logit-based model. Several observations can be made from the table. First, in the passenger context, a wide variety of modes are considered. However, only truck and rail modes are considered in the freight context. The focus in these studies is the competition (and the potential trade-off) between rail and truck mode. Second, in the passenger and freight related studies, Artificial Neural Network (ANN) and its variants are the most commonly used ML classifier used to examine mode choice. The ML classifiers are compared against utility-based logit/probit models such binary logit/probit and MNL. The data sources include both revealed preference and stated preference surveys. Third, the majority of the studies used accuracy (success rate), precision, recall and confusion matrix to compare the performances of the classical and ML classifier methods. Fourth, results from past research are somewhat mixed in comparing the performance of traditional discrete choice models and ML classifiers; the majority of the studies finding that ML methods are better while a few reports result on the contrary. As expected, depending on the data types (i.e., data from a single survey or data compiled from multiple sources), performance of the models was found to be different.

[Table 1 near here]

While earlier research has contributed substantially to our understanding of factors influencing mode choice, several gaps still exist. There is a paucity of research on the examination of freight mode choice using ML methods; our review yielded only four studies (Abdelwahab and Sayed, 1999; Sayed and Razavi, 2000; Sayed et al., 2003; Tortum et al., 2009). Their analysis efforts are limited to only ANN and neuro fuzzy approaches and no conclusive evidence regarding improvement in predictive accuracy over classical models was reported. In the passenger mode choice context, earlier research has highlighted improvements in prediction accuracy with ML approaches. However, the performance of other ML methods is yet to be investigated in freight mode choice analysis.

The current research is motivated by the need to undertake a detailed comparison of the widely used ML classifier methods for modeling fright mode choice. Using the 2012 Commodity Flow Survey (CFS) data, we investigate the predictive performance of a host of widely used ML classifier methods—Naïve Bayes (NB), Support Vector Machine (SVM), ANN, K-Nearest Neighbors (KNN), Classification and Regression Tree (CART), Random Forest (RF), Boosting (BOOST) and Bagging (BAG)—with the traditional MNL model. Previous freight-related studies using ML methods only considered two modes for analysis—truck and rail. In our analysis, we consider 5 modes including truck, rail, air, water and parcel. Moreover, the analysis is not confined to using variables only available in the dataset; we augmented it with spatial attributes (several origin-destination attributes and CFS zonal level variables) from secondary data sources. Finally, the current research also examines the role of sample size and training-testing data split ratios on the predictive ability of the various approaches.

**Data**
The primary data source used in this study is the 2012 Commodity Flow Survey (CFS). This data is publicly available and has a total of 4,547,661 shipment records from approximately 60,000 responding businesses and industries. Each shipment record includes some important freight characteristics (e.g., shipment size, value, distance, commodity type and origin and destination CFS area). The CFS reports 21 modes of transport. Since many of these alternatives have insignificant sample share, the reported modes were consolidated into five major groups: (1) for-hire truck, (2) private truck, (3) parcel service, (4) air and (5) other mode. For-hire truck mode represents the trucks operated by non-governmental businesses to provide services to customers under a negotiated rate. Private truck mode represents trucks owned and used by individual businesses for their own freight movement. Parcel service refers to a combination of modes (on ground/air/express carrier). Air mode consists of both air and truck since truck is needed to pick up and/or deliver the shipment from origin and/or destination which cannot be accessed by air. Lastly, the other mode consists of rail, water, pipeline or combination of non-parcel multiple modes. The weighted mode shares are for-hire truck (16.58%), private truck (26.06%), parcel service (55.85%), air (1.36%) and other (0.16%).

Shipment size, value and distance are reported as continuous variable in the CFS data. Shipment size and value were categorized into seven and five groups, respectively. These groups were created by making sure that each group has reasonable sample share. The groups for shipment size are $\leq 30$, 31–200, 201–1000, 1001–5000, 5001–30000, 3001–45000 and >45000 lb. The groups for shipment value are <$300, $300–$1000, $1001–$5000 and >$5000. Shipment distance was categorized into eight groups following the Freight Analysis Framework distance band (Freight Analysis Framework, 2019). The groups are <100, 100–249, 250–499, 500–749, 750–999, 1000–1499, 1500–200 and >2000 mile. The other freight characteristics include Standard Classification of Transported Goods (SCTG) commodity types, hazardous materials, temperature-controlled, export and origin and destination CFS area. For SCTG commodity types, based on the information provided in the CFS, shipments were consolidated into nine groups. Additionally, a number of origin-destination (OD) attributes and CFS zonal level (both at origin and destination) variables were generated utilizing several secondary data sources. The details of these variables and sources can be found in Keya et al. (2019). In this study, only those secondary variables found as statistically significant in freight mode choice in the above study is used. The resulting dataset has 21 variables, described in Table 2.

[Table 2 near here]

**Methods Description**

*Classifiers*
In this section we describe the methodologies used in the analysis. For the sake of brevity, we only provide the conceptual underpinning of the methods. In addition, we also mention how we calibrated and adapted the methods for our study.

In our study, classical MNL is considered as the baseline classifier. Choice of mode is unordered; hence, MNL is the widely used method to examine it. In the MNL model, the probabilities describing the possible outcomes of observation are modeled using a logistic function. The model is implemented as a classifier in which an optimization problem is solved that minimizes a cost function (Flach, 2012).

NB is a supervised learning method based on Bayes theorem with the assumption of conditional independence between every pair of features. To calculate probabilities from features, their probability distributions are estimated. In this study, Gaussian NB is used; that is, the probability of the features is assumed to be Gaussian. It is used because some of the input variables are real-valued. NB requires a small amount of training data to estimate the necessary parameters (Flach, 2012).

SVM is a supervised learning method with associated learning algorithms that analyze data used for classification. It classifies observations by projecting the variables into a high-dimensional feature space. The class that receives the most votes from all classifiers is chosen during prediction. In this study, an SVM with a linear kernel is specified for the decision function. Linear kernel provides more flexibility in the choice of penalties and loss functions and scales better to a large number of samples (Flach, 2012).

The ANN method used here is the Multi-layer Perceptron (MLP), which is a supervised learning algorithm and is inspired by biological neural networks. Given a set of features and a target, it can learn a non-linear function approximation for classification. Typically, the classifier with one hidden layer is trained using Backpropagation methods by minimizing the Cross-Entropy loss function (Flach, 2012). In this study, an ANN with a single hidden layer of 100 neurons is used.

KNN is an unsupervised learning algorithm based on the *k* nearest neighbors of each query point. It finds *k* number of training samples closest in distance to the new point and predicts the label from these. Classification is computed from a simple majority vote of the nearest neighbors of each point (Flach, 2012). In this study, a KNN with 5 neighbors is used.

CART is a non-parametric supervised learning method having a tree-like data structure. It creates a model that predicts the value of a target variable by learning simple decision rules inferred from the data. The nodes of the tree represent decision rules which split the feature space and the leaves of the tree represent the classes. CART is simple to understand and to interpret and requires little data preparation. However, CART learners create biased trees if some classes dominate (Flach, 2012).

RF is an ensemble meta estimator that fits a number of CART classifiers on various sub-samples of the dataset. In other words, instead of fitting a single "best" tree model, the RF strategically combines multiple simple decision trees to optimize predictive performance. Each tree in the ensemble is built from a sample drawn with replacement from the training set. Also, when splitting a node during the construction of the tree, the split that is picked is the best split among a random subset of the features. RF uses averaging to improve the predictive accuracy and control over-fitting (Flach, 2012). In this study, an RF consisting of 10 trees is used and two randomly selected variables are considered for each split at the tree nodes.

BOOST is an ensemble meta estimator that fits a sequence of weak leaners on repeatedly modified versions of the data. The predictions from all of them are then combined through a weighted majority vote to produce the final prediction. Here the gradient boosting model is used, where in each stage a number of trees are fit on the negative gradient of the multinomial deviance loss function (Flach, 2012). In this study, 100 trees are fitted in total.

BAG is an ensemble meta estimator that fits base classifiers on random subsets of the original dataset. Then the final prediction is formed by aggregating the individual predictions (either by voting or by averaging). It works best with strong and complex models (e.g., fully developed CART) (Flach, 2012). In this study, 10 classification trees are bagged.

*Model Comparison*

To compare the performance of the classifiers, we use a combination of k-fold cross validation and holdout method. First, using a split ratio (varied from 0.1 to 0.6), the dataset is partitioned into training and testing datasets. Then, the training dataset is evaluated utilizing k-fold cross validation. This procedure randomly partitions the training data into k disjoint subsets. One subset at a time is then used for testing the model, while k−1 sets are used to build the model. For different k values and split ratio, the classifiers are evaluated. Next, based on the entire training dataset, the models are estimated. These models are then applied to the testing dataset for prediction purposes.

The classification performance of the models is evaluated using four metrics: accuracy, precision, recall and F1-score. Accuracy describes the percentage of correctly classified observations in a dataset; it is computed as the number of correctly classified observations, divided by the total number of observations classified. Precision measures the proportion of correctly classified observations among all of those observations that were similarly classified; it is computed as the number of correctly classified observations of a particular mode, divided by the total number of observations classified as that mode. Recall measures the proportion of observations of a particular class that are correctly classified; it is computed as the number correctly classified observations of a particular mode, divided by the total number of actual mode observations in the dataset. F1-score is the harmonic mean of precision and recall. Let us assume we have two modes, namely A and B, and a total of 100 observations. After prediction, it is found that the machine learning method correctly classified 40 mode A observations as mode A, 20 mode A observations as mode B, 10 mode B observations as mode A and 30 mode B observations as mode B. The accuracy of the classifier is 70/100 = 0.7; precision for mode A is 40/50 = 0.8; recall for mode A is 40/60 = 0.67; and F1-score for mode A is 2×(0.8×0.67)/(0.8+0.67) = 0.73.

**Results and Discussion**

All modeling and analysis were completed in Python and using the "scikit-learn" package (Pedregosa et al., 2011). This package provides efficient tools for machine learning classification methods. A standard desktop computer was used to run all the experiments.

The analysis was conducted in the following steps. First, a sample was drawn for mode share analysis from CFS database. Second, for three instances of cross validation (10-, 20- and 30-fold) and six instances of testing-training data split ratio (0.1, 0.2, 0.3, 0.4, 0.5 and 0.6), different classifiers were estimated. The various performance metrics described above were evaluated for all classifiers. Finally, the process was repeated with different sample sizes—136,073 records, 226,785 records, 453,574 records, 907,139 records, 1,360,706 records, 1,814,271 records and 2,267,842 records—to illustrate the influence of sample size on the classifier performance.

While presenting detailed results for all scenarios considered is beyond the scope of the paper, we illustrate the exercise by presenting detailed results for one sample size (136,073 records). Figure 1 presents the results of these experiments and compares the classifiers. For instance, Figure 1(a) shows the accuracy results for 0.1 split ratio (i.e., 90% training and 10% test data) under three cross validation sizes. For 10-fold cross validation, with respect to mean accuracy, RF had the best results (0.754), closely followed by BAG (0.750) and BOOST (0.749). The next best classifiers were CART (0.711), NB (0.672) and KNN (0.617). The accuracy for ANN (0.506) was slightly higher than MNL (0.420). SVM had the lowest accuracy for all classifiers with 0.367. In addition, SVM and ANN had very wide range in accuracy values (i.e., high standard deviation). For both 20- and 30-fold cross validation, similar trends were observed. Comparing the results over cross validation sizes, it is generally observed that the range of accuracy values becomes wider with a higher number of cross validations for all classifiers. The reader would note that, to fit the models and to get predictions from test data, sample weights provided in the dataset were used.

Comparing different data split ratio results, a similar pattern in accuracy values, as described above, was evident. Based on the results, the optimal data split ratio can be obtained, which is the one with the smallest standard deviation. This optimal ratio helps to build efficient models and to make better predictions. Considering all six data split ratio results, shown in Figure 1, it is found that the accuracy values had the smallest standard deviation with ratio 0.3 for most of the classifiers. In addition, they had higher median accuracy at this split ratio. For that reason, we conclude that split ratio of 0.3 is the optimal ratio (70% training and 30% test). This ratio split was used for the subsequent experiments.

[Figure 1 near here]

The analysis was repeated for the 6 other samples identified earlier. The results of the experiments using the above seven samples and holdout method (ratio 0.3) are provided in Figure 2 and Table 3. Figure 2 presents the accuracy results of the classifiers under different sample sizes. On the other hand, Table 3 presents the precision, recall and F1-score for each mode under RF, BAG and BOOST classifiers. The accuracy for RF, BAG, CART and KNN increased or remained the same with the increase in sample size while the accuracy for NB decreased with the increase in sample size. There is no distinct pattern present in accuracy values for ANN, MNL and SVM. Furthermore, for SVM, accuracy values substantially varied with the sample size. Several conclusions can be made from the above results. For freight mode choice modeling, the tree-based ensemble classifiers performed the best. However, increasing the sample size is not always associated with increased accuracy. If the accuracy increases, the amount is not that high (~1%). Given that the accuracy of MNL is not that high, it is recommended to use machine learning classifiers if sufficient observations are available.

[Figure 2 near here]
[Table 3 near here]

From the various approaches, the three methods that offered the highest accuracy were considered for further analysis using precision, recall and F1-score. The measures by mode are presented for RF, BAG and BOOST methods in Table 3 for different sample sizes. The results highlight how increasing sample sizes contribute to improved recall values. However, for modes with smaller share such as Air, increasing sample size does not necessarily increase recall values (see BAG and BOOST classifier accuracy for air mode for different sample sizes). For modes with larger share such as truck modes, there is a general improvement in recall with sample size (while tapering is observed at the larger end of sample sizes). Similar relationships are observed for Precision and F1-scores. Overall, RF, BAG and BOOST performed well based on the estimates.

To further offer insights on the variable affecting classifier performance, we plot importance of each variable for RF and BOOST classifiers in Figure 3. The importance values were obtained from experiments performed using different sample sizes. For RF, shipment distance is found to be the most important variable. The other variables with high importance are shipment size, industry classification of shipper and SCTG commodity type. Interestingly, shipment value is found to have moderate importance. This finding contradicts Abdelwahab and Sayed (1999) where "shipment value" was found not to have significant impact in the ANN model. Note that the other variables with moderate importance are origin CFS area, destination CFS area, density of employees at origin, number of warehouse and super center at origin, density of highway at both origin and destination, density of railway at both origin and destination, and population density at destination. For BOOST, industry classification is found to be the most important variable. Similar to RF results, shipment distance, size, and commodity type have high importance.

Overall, shipment characteristics have higher importance, and OD attributes and CFS zonal level variables have moderate to low importance in classifying freight mode choice. The result related to shipment characteristics being more important for freight mode choice corroborates other related studies (Sayed et al., 2003; Abdelwahab and Sayed, 1999). Since some of the OD attributes have moderate importance in both classifiers, it is recommended to use these attributes in addition to the shipment characteristics for modeling freight mode choice.

[Figure 3 near here]

**Conclusion and Future Research**
Efficient and cost-effective freight movement is a prerequisite to a region's economic viability, growth, prosperity, and livability. The mode chosen for freight transportation has significant implications for the transportation system and the environment at large. Traditionally, the research modeling freight mode choice has typically focused on the random utility based multinomial logit (MNL) model and its variants. Over the last decade, driven by the enhancements in computing power and the advent of big data analytics, machine learning approaches have gained attention from transportation researchers. With their inherent strength in handling large datasets, these approaches are well suited to extracting patterns that are often hard to accommodate within traditional econometric models. Despite the advantages, there is a paucity of research on the examination of freight mode choice using machine learning methods. The few studies found are limited in their number of alternatives consideration, number of observations used, and number of ML classifiers used. In our research, we aim to address these gaps in the literature.

More specifically, we investigated the predictive performance of a host of widely used machine learning classifier methods—Naïve Bayes (NB), Support Vector Machine (SVM), Artificial Neural Network (ANN), K-Nearest Neighbors (KNN), Classification and Regression Tree (CART), Random Forest (RF), Boosting (BOOST) and Bagging (BAG)—with the traditional multinomial logit (MNL) model. The 2012 Commodity Flow Survey (CFS) data augmented with several spatial attributes (origin-destination attributes and CFS zonal level variables) from secondary data sources were used. A combination of k-fold cross validation and holdout method were employed. The major findings include: (a) among the investigated classifiers, RF produced the most accurate predictions, closely followed by BOOST and BAG, (b) the performance of MNL model was lower than all other classifiers, except SVM, (c) air mode had the lowest prediction accuracy among all classifiers. Furthermore, private trucks had higher prediction than for-hire truck for all classifiers, except MNL and SVM and (d) the top three important variables for freight mode choice were shipment distance, industry classification of the shipper and shipment size.

In future, the authors would like to incorporate several level of service measures (e.g., shipping cost, operating cost, shipping time) to the variables list to investigate if they have any impact on the freight mode choice decisions using the above machine learning classifiers.

**Acknowledgment**
The authors would like to acknowledge Nowreen Keya and Naveen Chandra for helping with initial data preparation.

**Author Contributions**
The authors confirm contribution to the paper as follows: study conception and design: S. Anowar, M. Uddin, N. Eluru; data collection: M. Uddin, S. Anowar; analysis and interpretation of results:

**References**

Abdelwahab, W., and T. Sayed. Freight Mode Choice Models using Artificial Neural Networks. Civil Engineering Systems, 1999. 16: 267–286.

Austin, D. Pricing Freight Transport to Account for External Costs. *Congressional Budget*, 2015.

Biagioni, J.P., P.M. Szczurek, P.C. Nelson, and A. Mohammadian. Tour-Based Mode Choice Modeling: Using an Ensemble of (Un-) Conditional Data-Mining Classifiers. Presented at the 88th Annual Meeting of the Transportation Research Board, Washington, D.C., 2009.

Cantarella, G.E., and S. de Luca. Multilayer Feedforward Networks for Transportation Mode Choice Analysis: An Analysis and A Comparison with Random Utility Models. Transportation Research Part C, 2005. 13: 121–155.

Cheng, L., X. Chen, J. De Vos, X. Lai, and F. Witlox. Applying A Random Forest Method Approach to Model Travel Mode Choice Behavior. Travel Behaviour and Society, 2019. 14: 1–10.

Flach, P. *Machine Learning: The Art and Science of Algorithms that Make Sense of Data*. Cambridge University Press, Cambridge, UK, 2012.

*Freight Analysis Framework*. Center for Transportation Analysis, Oak Ridge National Laboratory. faf.ornl.gov/fafweb/. Accessed May 15, 2019.

Golob, T.F., and A.C. Regan. Impacts of Information Technology on Personal Travel and Commercial Vehicle Operations: Research Challenges and Opportunities. *Transportation Research Part C*, 2001. 9: 87–121.

Hagenauer, J., and M. Helbich. A Comparative Study of Machine Learning Classifiers for Modeling Travel Mode Choice. *Expert Systems with Applications*, 2017. 78: 273–282.

Hensher, D.A., and T.T. Ton. A Comparison of the Predictive Potential of Artificial Neural Networks and Nested Logit Models for Commuter Mode Choice. *Transportation Research Part E*, 2000. 36: 155–172.

Huang, H., M. Abdel-Aty, and A.L. Darwiche. County-Level Crash Risk Analysis in Florida: Bayesian Spatial Modeling. *Transportation Research Record,* 2010. 2148: 27–37.

Keya, N., S. Anowar, and N. Eluru. Estimating a Freight Mode Choice Model: A Case Study of Commodity Flow Survey 2012. Presented at the 96th Annual Meeting of the Transportation Research Board, Washington, D.C., 2017.

Keya, N., S. Anowar, and N. Eluru. Joint Model of Freight Mode Choice and Shipment Size: A Copula-Based Random Regret Minimization Framework. *Transportation Research Part E*, 2019. 125: 97–115.

Li, J., J. Weng, C. Shao, and H. Guo. Cluster-Based Logistic Regression Model for Holiday Travel Mode Choice. *Procedia Engineering*, 2016. 137: 729–737.

Lindner, A., C.S. Pitombo, and A.L. Cunha. Estimating Motorized Travel Mode Choice using Classifiers: An Application for High-Dimensional Multicollinear Data. *Travel Behaviour and Society*, 2017. 6: 100–109.

Moons, E., G. Wets, and M. Aerts. Nonlinear Models for Determining Mode Choice. In *Lecture Notes in Artificial Intelligence*, edited by Jose Neves, Manuel Filipe Santos, and Jose Manuel Machado, 2007. 183–94.

Nam, D., H. Kim, J. Cho, and R. Jayakrishnan. A Model Based on Deep Learning for Predicting Travel Mode Choice. Presented at the 96th Annual Meeting of the Transportation Research Board, Washington, D.C., 2017.

Nijkamp, P., A. Reggiani, and T. Tritapepe. Modelling Inter-Urban Transport Flows in Italy: A Comparison between Neural Network Analysis and Logit Analysis. *Transportation Research Part C*, 1996. 4: 323–338.

Omrani, H. Predicting Travel Mode of Individuals by Machine Learning. *Transportation Research Procedia*, 2015. 10: 840–849.

Omrani, H., O. Charif, P. Gerber, A. Awasthi, and P. Trigano. Prediction of Individual Travel Mode with Evidential Neural Network Model. *Transportation Research Record: Journal of the Transportation Research Board*, 2013. 2399: 1–8.

Pedregosa, F., G. Varoquaux, A. Gramfort, et al. Scikit-learn: Machine Learning in Python. *Journal of Machine Learning Research*, 2011. 12: 2825–2830.

Ponnuswamy, S, and D. Johnson Victor. *Urban Transportation: Planning, Operation and Management*. New York: McGraw-Hill Education, 2012.

Pulugurta, S., A. Arun, and M. Errampalli. Use of Artificial Intelligence for Mode Choice Analysis and Comparison with Traditional Multinomial Logit Model. *Procedia - Social and Behavioral Sciences*, 2013. 104: 583–592.

Rotem-Mindali, O., and J.W.J. Weltevreden. Transport Effects of e-Commerce: What can be Learned After Years of Research? *Transportation*, 2013. 40: 867–885.

Sayed, T., and A. Razavi. Comparison of Neural and Conventional Approaches to Mode Choice Analysis. *Journal of Computing in Civil Engineering*, 2000. 14:23–30.

Sayed, T., A. Tavakolie, and A. Razavi. Comparison of Adaptive Network Based Fuzzy Inference Systems and B-spline Neuro-Fuzzy Mode Choice Models. *Journal of Computing in Civil Engineering*, 2003. 17: 123–130.

Sekhar, C.R., Minal, and E. Madhu. Mode Choice Analysis Using Random Forrest Decision Trees. *Transportation Research Procedia*, 2016. 17: 644–652.

Tortum, A., N. Yayla, and M. Gökdağ. The Modeling of Mode Choices of Intercity Freight Transportation with the Artificial Neural Networks and Adaptive Neuro-Fuzzy Inference System. *Expert Systems with Applications*, 2009. 36: 6199–6217.

Wets, G., K. Vanhoof, T. Arentze, and H. Timmermans. Identifying Decision Structures Underlying Activity Patterns: An Exploration of Data Mining Algorithms. *Transportation Research Record: Journal of the Transportation Research Board*, 2000. 1718: 1–9.

Xie, C., J. Lu, and E. Parkany. Work Travel Mode Choice Modeling with Data Mining: Decision Trees and Neural Networks. *Transportation Research Record: Journal of the Transportation Research Board*, 2003. 1854: 50–61.

Zhang, Y., and Y. Xie. Travel Mode Choice Modeling with Support Vector Machines. *Transportation Research Record: Journal of the Transportation Research Board*, 2008. 2076: 141–150.

Zhu, Z., X. Chen, C. Xiong, and L. Zhang. A Mixed Bayesian Network for Two-Dimensional Decision Modeling of Departure Time and Mode Choice. *Transportation*, 2018. 45: 1499–1522.

**List of Figures**

**List of Tables**

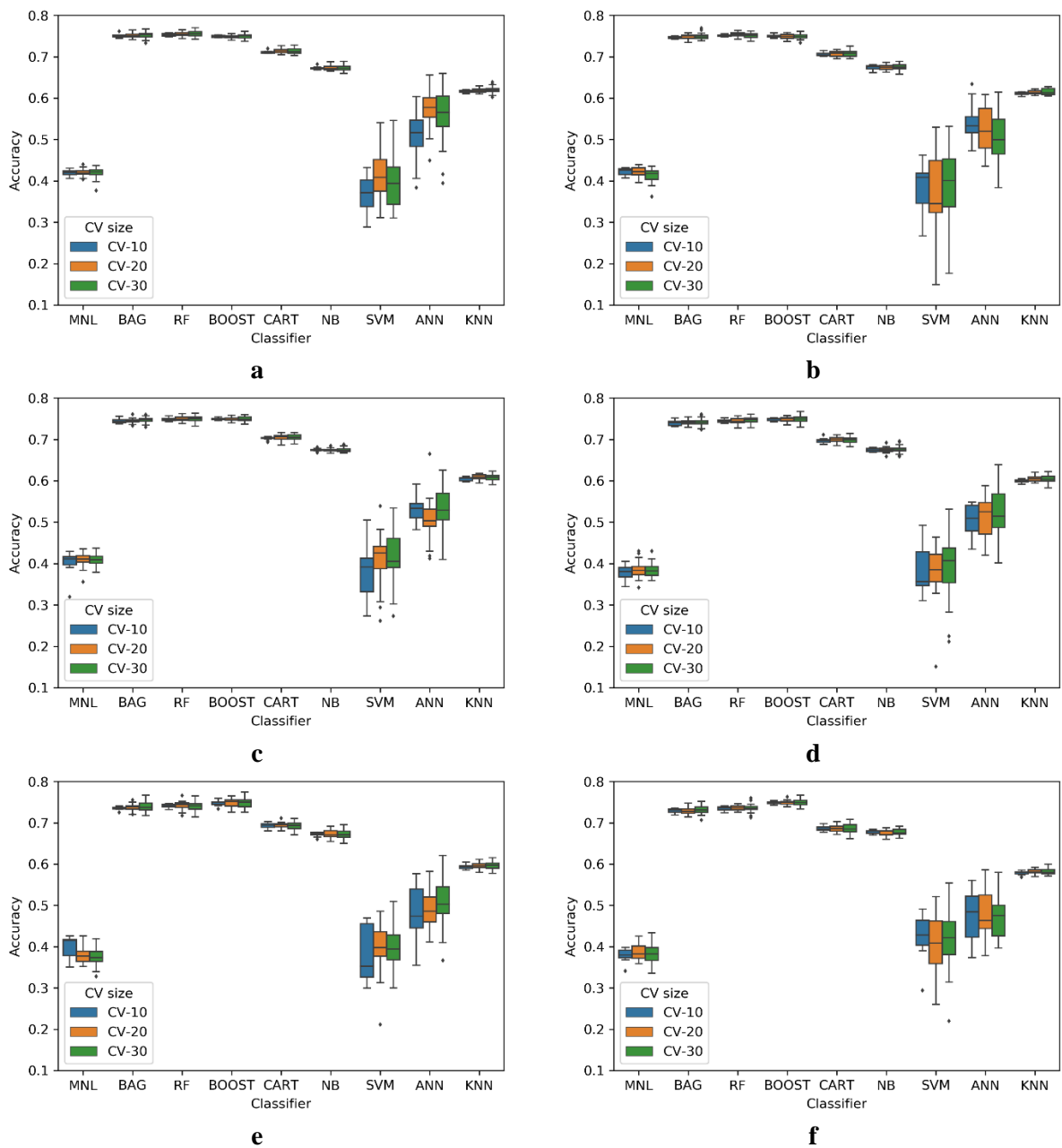

**FIGURE 1** Comparison of classifiers under three cross validation sizes and different testing-training data split ratios: (a) 0.1, (b) 0.2, (c) 0.3, (d) 0.4, (e) 0.5 and (f) 0.6

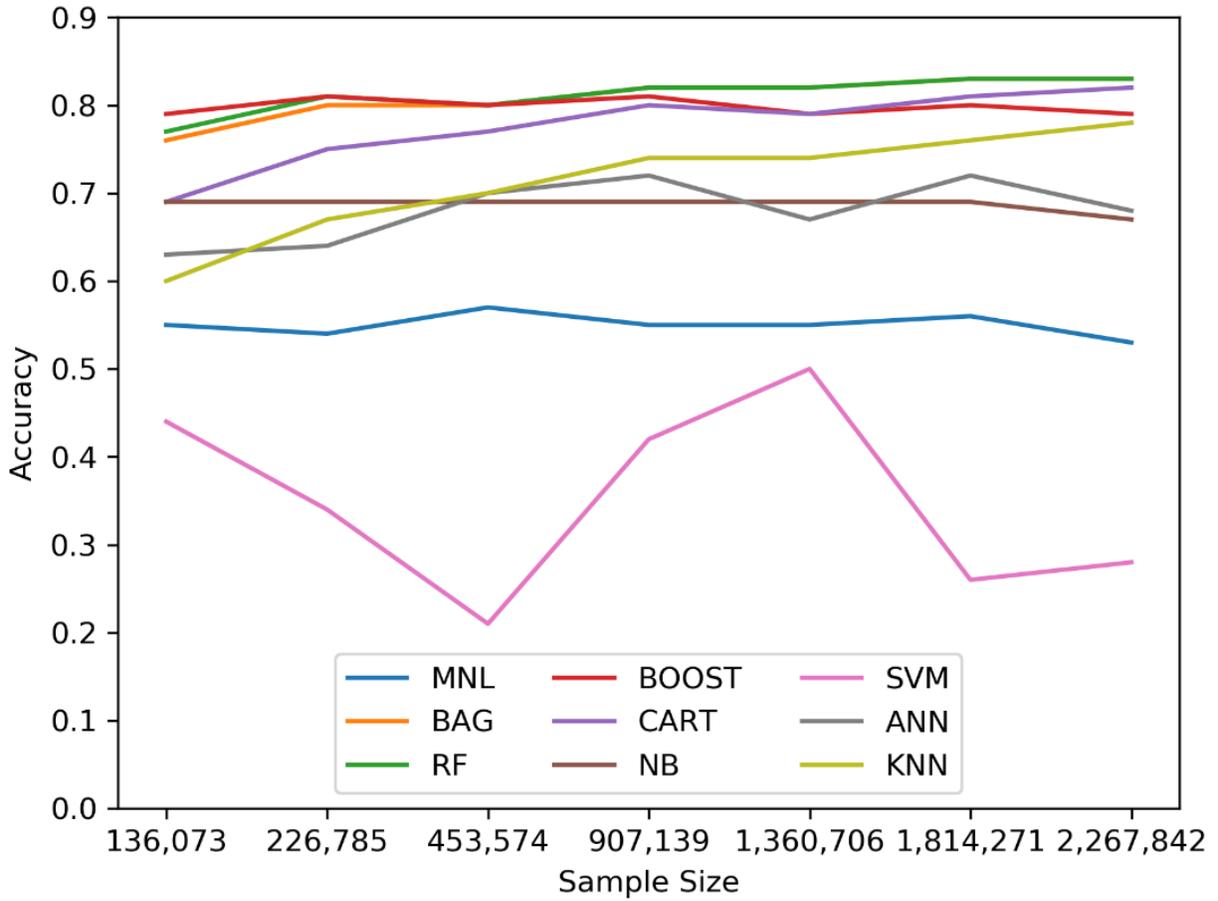

**FIGURE 2 Comparison of classifiers under different samples sizes**

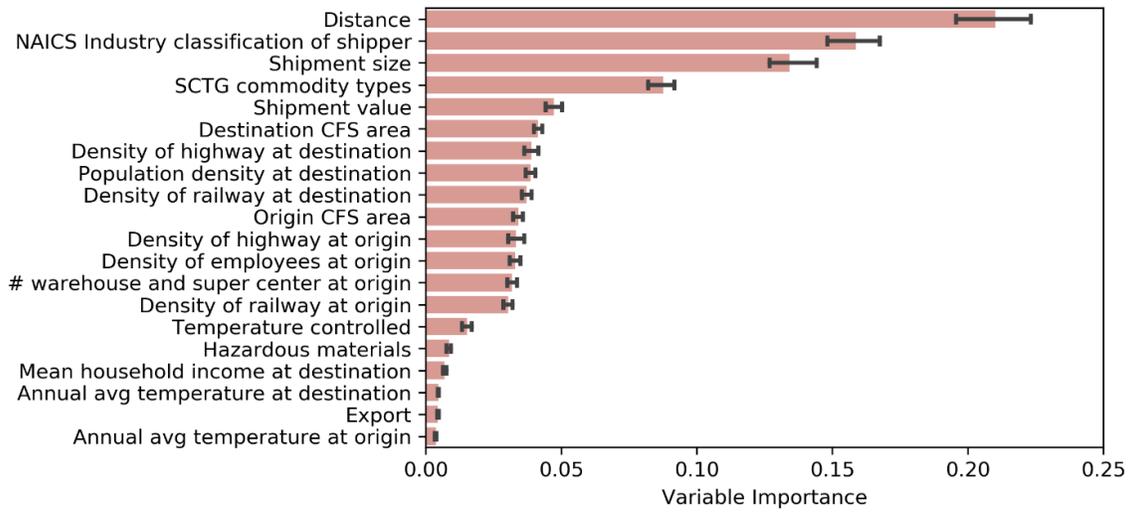

(a)

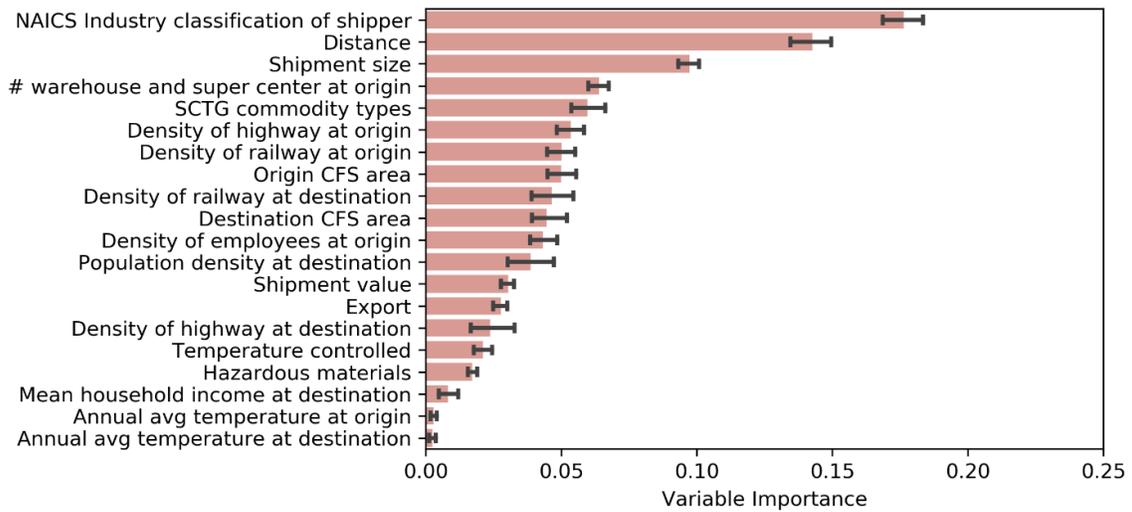

(b)

**FIGURE 3 Variable importance: (a) RF and (b) BOOST**

**TABLE 1 Previous literature on the use of machine learning for passenger and freight mode choice**

| Study | Study Area | Data Source | Mode(s) Considered | Machine Learning Methods Used | Classical Methods Used | Comparison Metrics (Level of Prediction) | Training and Testing Data | Findings |
|---|---|---|---|---|---|---|---|---|
| | | | *Passenger* | | | | | |
| Nijkamp et al. (1996) | Italy | Census data | Rail, road | FNN | BL | Absolute residuals; relative residuals (Aggregate) | Training (698); Test (349) | • Both ANN and logit models provide good performance; ANN is marginally better |
| Hensher and Ton (2000) | Sydney and Melbourne, Australia | Commute mode choice survey (stated preference) | Drive alone, ride share, bus, busway, train, light rail | ANN | NL | Prediction success (Aggregate) | - | • No clear indication of which model is better |
| Wets et al. (2000) | South Rotterdam, The Netherlands | Activity survey | Car driver, car passenger, bike, walk, public transport | C4 Algorithm | CHAID; MNL | Predicted mode share (Aggregate) | Training (3374); Test (1124) | • None of the models clearly outperforms the other |
| Xie et al. (2003) | San Francisco, USA | San Francisco Bay Area Travel Survey | SOV (car, van, motorcycle, moped), carpool, transit, bike, walk | DT; ANN-MLP | MNL | Confusion matrix; Correctly predicted mode share (Aggregate and disaggregate) | Training (2373); Test (2373) | • All three models provide comparable prediction performance |
| Cantarella and de Luca (2005) | Veneto, Italy | Traveler interview survey (stated preference) | Car, carpool, bus | MFFN | MNL; NL; CNL | Predicted mode share difference; MSE (Aggregate) RMSE, Fitting factor (Disaggregate) | Training (23%); Test (77%) | • MLFFN outperforms classical models |
| Moons et al. (2007) | South Rotterdam, The Netherlands | Activity survey | Car driver, car passenger, bike, walk, public transport | SVM; CART | MNL; MFS; | % of correct prediction (Aggregate) | Training (70%); Test (30%) | • On skewed dataset, classical models perform better; on balanced dataset, machine learning methods perform better |
| Biagioni et al. (2009) | Chicago, USA | Chicago Travel Tracker Survey | Walk, bike, auto-drive, auto-passenger, bus, train, Pace bus, commuter rail | DT; NB; simple logistic; SVM | MNL | Accuracy, precision, recall (Aggregate) | - | • NB and DT performed the best but the performance of MNL was reasonable |

| Study | Study Area | Data Source | Mode(s) Considered | Machine Learning Methods Used | Classical Methods Used | Comparison Metrics (Level of Prediction) | Training and Testing Data | Findings |
|---|---|---|---|---|---|---|---|---|
| Zhang and Xie (2008) | San Francisco Bay Area, USA | Commute trip data | Drive alone, shared ride, transit, bike, walk | SVM; MFNN | MNL | Confusion matrix (Aggregate) | Training (75%); Testing (25%) | • SVM outperforms MNL |
| Pulugurta et al. (2013) | Port Blair, India | Household travel survey | Car, bus, two-wheeler, three-wheeled auto-rickshaw, cycle, cycle rickshaw, walk | Sugeno-type fuzzy logic model | MNL | Classification accuracy; predicted mode share (Aggregate) | Training (85); Testing (15%) | • Prediction accuracy found: MNL (40%); fuzzy logit model (70%)<br>• Predicted mode share by fuzzy logic model is closer to the actual values |
| Omrani et al. (2013) | Luxembourg City, Luxembourg | Socioeconomic Panel Survey | Car, public transit, walk/bike | ENN; ANN – MLP; ANN -RBF; DT; Bayes; K-NN; SVM | MNL | Confusion matrix; success rate; (Aggregate) ternary plot (Disaggregate) | Training (60); Testing (40) | • Success rate found: MNL (62%); ANN (80-81%); ENN (83%); DT (78%); Bayes (67%); $k$-NN (77%); SVM (80%) |
| Sekhar et al. (2016) | Delhi, India | Household survey | Car, carpool, two-wheeler, bus, metro, three-wheeler, bicycle, walk | DT; RF | MNL | Prediction accuracy (Aggregate) | - | • Prediction accuracy found: MNL (77%); RF (99%) |
| Omrani (2015) | Luxembourg City, Luxembourg | Socioeconomic Panel Survey | Car, Public transit, Walk/bike | ANN-MLP; ANN-RBF; SVM | MNL | Confusion matrix; success rate; (Aggregate) ternary plot (Disaggregate) | Training (60); Testing (40) | • Percentage of correct prediction found: MNL (65%); SVM (68%); ANN-RBF (80%); ANN-MLP (82%) |
| Li et al. (2016) | Beijing, China | Holiday travel data | Car, non-car | Cluster based logistic regression | BL | Prediction accuracy (Aggregate) | - | • Cluster based Logistic regression performed better |
| Hagenauer and Helbich (2017) | The Netherland | Dutch National Travel Survey | Walk, bike, car, public transit | NB; SVM; ANN; DT; Boosting; Bagging; RF | MNL | Prediction accuracy; sensitivity statistics (Aggregate) | 10-fold cross validation | • RF performs better than MNL |
| Lindner et al. (2017) | Sao Paulo, Brazil | O-D Survey | Car/Motorcycle; Public transit | CT; ANN | BL | Average prediction accuracy | Training (70); Testing (30) | • Percentage of correct prediction found: BL (74%); ANN-MLP (79%); CT (80%) |

| Study | Study Area | Data Source | Mode(s) Considered | Machine Learning Methods Used | Classical Methods Used | Comparison Metrics (Level of Prediction) | Training and Testing Data | Findings |
|---|---|---|---|---|---|---|---|---|
| Nam et al. (2017) | Switzerland | Mode choice survey of long-distance travel (stated preference) | Car, rail. Metro | ANN; ANN – MLP; DNN | NL; CNL | Log-likelihood (Disaggregate) % of correct predictions (Aggregate) | - | • Percentage of correct prediction found: NL (64%); CNL (64%); ANN-MLP (59-63%); DNN (65-67%) |
| Zhu et al. (2018) | Washington and Baltimore, USA | Household travel survey | Public transit, car driver, car passenger, walk/bike | BN; DT | NL | Match rate (Disaggregate) | 5 subsets for cross-validation | • Success rate found: NL (~47%); BN (45-47%); DT (44-45%)) |
| Cheng et al. (2019) | Nanjing, China | Household travel suvey | Walk, bicycle, e-motorcycle, public transport, automobile | RF; SVM; Boosting | MNL | Accuracy; MAPE | Training (80%); Testing (20%) | • RF and SVM provide the best prediction accuracy |
| *Freight* | | | | | | | | |
| Abdelwahab and Sayed (1999) | USA | Commodity Transportation Survey | Rail, truck | ANN | BL; BP | Success rate (Aggregate) | Training (1,000); Testing (586) | • ANN provides equal or higher predictive accuracies |
| Sayed and Razavi (2000) | USA | Commodity Transportation Survey | Rail, truck | ANN; Neuro Fuzzy approach (Adaptive B-spline networks) | BL | Success rate (Aggregate) | Training (5,000); Testing (2500) | • Neuro Fuzzy uses fewer variables than others to achieve the same predictive accuracy of the other models |
| Sayed et al. (2003) | USA | Commodity Transportation Survey | Rail, truck | ANFIS; B-Spline associative memory networks | --- | Success rate (Aggregate) | Training (5,000); Testing (2500) | • B-Spline AMN require fewer variables to achieve the same performance |
| Tortum et. al. (2009) | Turkey, Germany, France, Austria | Freight flow data compiled from multiple sources | Rail, truck | ANFIS; ANN-MLP | LR; MRM | (Aggregate) | Training (80); Testing (20%) | • ANN and ANFIS models perform better than the classical models |

* ANFIS = Adaptive Neuro Fuzzy Inference System; ANN-MLP = Artificial Neural Network - Multi-Layer Perceptron; ANN-RBF = Artificial Neural Network - Radial Basis Function; BL = Binary Logit; BN = Bayesian Network; BP = Binary Probit; CART = Classification and Regression Tree; CNL = Cross nested logit; CT = Classification Tree; DNN = Deep Neural Network; DT = Decision Tree; ENN = Evidential Neural Network; FNN = Feedforward Neural Network; *k*-NN = K-Nearest Neighbors; LR = Linear Regression; MAPE = Mean Absolute Percent Error; MR = Multiple Regression; MSE = Mean Square Error; NB = Naïve Bayes; NL = Nested logit; O-D = Origin-Destination; RF = Random Forrest; RMSE = Root Mean Square Error; SOV = Single Occupant Vehicle; SVM = Support Vector Machine

**TABLE 2** Description of the variables

| Variable | Description |
| --- | --- |
| Mode | Shipment mode (for-hire truck, private truck, parcel service, air, and other) |
| Size | Weight of shipment in lbs (≤30, 31–200, 201–1000, 1001–5000, 500–30000, 30001–45000, >45000) |
| Value | Value of shipment in $ (<300, 300-1000, 1001-5000, >5000) |
| Distance | Routed distance between shipment origin and destination in miles (<100, 100–249, 250–499, 500–749, 750–999, 1000–1499, 1500–2000, >2000) |
| Commodity type | SCTG commodity group (raw food; prepared products; stone and non-metallic minerals; petroleum and coal; chemical products; wood, papers and textiles; metals and machinery; electronics; and furniture and others) |
| HAZMAT | Hazardous materials (Class 3.0 Hazmat, other hazmat, and not hazmat) |
| Temperature-controlled | Temperature controlled shipment (yes, no) |
| Export | Export shipment (yes, no) |
| Origin CFS | CFS area of shipment origin (132 areas) |
| Destination CFS | CFS area of shipment destination (132 areas) |
| NAICS | Industry classification of shipper (45 classes) |
| Origin employee | Density of employees at origin |
| Origin warehouse | Number of warehouse and super center at origin |
| Origin highway | Density of highway at origin (mile/mile$^2$) |
| Origin railway | Density of railway at origin (mile/mile$^2$) |
| Origin temperature | Average temperature at origin in °F (≤60, >60) |
| Destination population | Population density at destination (1000 pop/mile$^2$) |
| Destination income | Mean household income at destination in $ (≥50,000, <50,000) |
| Destination temperature | Average temperature at destination in °F (≤60, >60) |
| Destination highway | Density of highway at destination (mile/mile$^2$) |
| Destination railway | Density of railway at destination (mile/mile$^2$) |

**TABLE 3 Performance of classifiers under different sample sizes**

| Sample Size | Freight Mode | RF | | | BAG | | | BOOST | | |
|---|---|---|---|---|---|---|---|---|---|---|
| | | Precision | Recall | F1-score | Precision | Recall | F1-score | Precision | Recall | F1-score |
| 136,073 | For-hire truck | 0.54 | 0.41 | 0.46 | 0.53 | 0.4 | 0.45 | 0.74 | 0.34 | 0.46 |
| | Private truck | 0.74 | 0.79 | 0.76 | 0.73 | 0.76 | 0.75 | 0.74 | 0.81 | 0.78 |
| | Parcel service | 0.83 | 0.88 | 0.85 | 0.83 | 0.89 | 0.86 | 0.82 | 0.93 | 0.87 |
| | Air | 0.14 | 0.04 | 0.07 | 0.13 | 0.05 | 0.08 | 0.54 | 0.05 | 0.09 |
| | Other mode | 0.66 | 0.3 | 0.42 | 0.54 | 0.38 | 0.45 | 0.7 | 0.2 | 0.31 |
| 226,785 | For-hire truck | 0.59 | 0.46 | 0.52 | 0.55 | 0.49 | 0.52 | 0.75 | 0.29 | 0.41 |
| | Private truck | 0.8 | 0.83 | 0.82 | 0.8 | 0.83 | 0.81 | 0.75 | 0.86 | 0.8 |
| | Parcel service | 0.86 | 0.92 | 0.89 | 0.87 | 0.89 | 0.88 | 0.84 | 0.94 | 0.89 |
| | Air | 0.36 | 0.09 | 0.14 | 0.36 | 0.15 | 0.21 | 0.32 | 0.06 | 0.1 |
| | Other mode | 0.64 | 0.29 | 0.4 | 0.59 | 0.34 | 0.43 | 0.63 | 0.22 | 0.32 |
| 453,574 | For-hire truck | 0.58 | 0.48 | 0.52 | 0.6 | 0.51 | 0.55 | 0.76 | 0.3 | 0.43 |
| | Private truck | 0.78 | 0.81 | 0.8 | 0.79 | 0.8 | 0.8 | 0.74 | 0.84 | 0.78 |
| | Parcel service | 0.85 | 0.9 | 0.87 | 0.86 | 0.9 | 0.88 | 0.83 | 0.94 | 0.88 |
| | Air | 0.28 | 0.09 | 0.14 | 0.35 | 0.1 | 0.15 | 0.19 | 0.02 | 0.04 |
| | Other mode | 0.77 | 0.4 | 0.53 | 0.67 | 0.44 | 0.53 | 0.56 | 0.23 | 0.32 |
| 907,139 | For-hire truck | 0.63 | 0.52 | 0.57 | 0.62 | 0.55 | 0.58 | 0.79 | 0.32 | 0.46 |
| | Private truck | 0.81 | 0.84 | 0.82 | 0.82 | 0.84 | 0.83 | 0.76 | 0.84 | 0.8 |
| | Parcel service | 0.86 | 0.91 | 0.89 | 0.87 | 0.9 | 0.89 | 0.83 | 0.95 | 0.89 |
| | Air | 0.39 | 0.13 | 0.19 | 0.39 | 0.14 | 0.2 | 0.4 | 0.02 | 0.04 |
| | Other mode | 0.74 | 0.49 | 0.59 | 0.74 | 0.55 | 0.63 | 0.61 | 0.3 | 0.4 |
| 1,360,706 | For-hire truck | 0.64 | 0.54 | 0.58 | 0.65 | 0.55 | 0.6 | 0.73 | 0.28 | 0.41 |
| | Private truck | 0.82 | 0.85 | 0.83 | 0.81 | 0.85 | 0.83 | 0.74 | 0.83 | 0.78 |
| | Parcel service | 0.87 | 0.91 | 0.89 | 0.88 | 0.9 | 0.89 | 0.82 | 0.94 | 0.87 |
| | Air | 0.35 | 0.16 | 0.22 | 0.25 | 0.19 | 0.22 | 0.48 | 0.07 | 0.12 |
| | Other mode | 0.77 | 0.5 | 0.61 | 0.76 | 0.55 | 0.63 | 0.69 | 0.25 | 0.37 |
| 1,814,271 | For-hire truck | 0.65 | 0.55 | 0.6 | 0.65 | 0.56 | 0.6 | 0.76 | 0.29 | 0.42 |
| | Private truck | 0.82 | 0.86 | 0.84 | 0.83 | 0.86 | 0.84 | 0.75 | 0.84 | 0.79 |
| | Parcel service | 0.88 | 0.91 | 0.9 | 0.88 | 0.91 | 0.9 | 0.83 | 0.95 | 0.88 |
| | Air | 0.33 | 0.15 | 0.21 | 0.42 | 0.17 | 0.25 | 0.45 | 0.05 | 0.1 |
| | Other mode | 0.76 | 0.55 | 0.64 | 0.75 | 0.56 | 0.64 | 0.55 | 0.24 | 0.33 |
| 2,267,842 | For-hire truck | 0.69 | 0.56 | 0.62 | 0.68 | 0.58 | 0.62 | 0.79 | 0.3 | 0.43 |
| | Private truck | 0.83 | 0.87 | 0.85 | 0.84 | 0.87 | 0.86 | 0.76 | 0.83 | 0.79 |
| | Parcel service | 0.87 | 0.91 | 0.89 | 0.88 | 0.91 | 0.89 | 0.81 | 0.95 | 0.87 |
| | Air | 0.5 | 0.23 | 0.32 | 0.42 | 0.26 | 0.32 | 0.52 | 0.05 | 0.1 |
| | Other mode | 0.78 | 0.58 | 0.66 | 0.75 | 0.6 | 0.67 | 0.58 | 0.24 | 0.34 |